\documentclass[letterpaper, 10 pt, conference]{ieeeconf}  

\IEEEoverridecommandlockouts                              

\usepackage{xcolor}

\title{RADIO-ViPE: Online Tightly Coupled Multi-Modal Fusion for Open-Vocabulary Semantic SLAM in Dynamic Environments}

\usepackage{tikz}
\def\checkmark{\tikz\fill[scale=0.4](0,.35) -- (.25,0) -- (1,.7) -- (.25,.15) -- cycle;} 

\usepackage{mathtools, nccmath}
\usepackage{amsmath}
\usepackage{amssymb}
\usepackage{algorithm}
\usepackage{algpseudocode}
\usepackage{bm}
\usepackage{graphicx}
\usepackage{booktabs}
\usepackage{authblk}
\usepackage{booktabs}
\usepackage{multirow}
\usepackage{colortbl}
\usepackage{xcolor}
\usepackage{graphicx}
 
\definecolor{bestcolor}{RGB}{180, 220, 170}       
\definecolor{secondcolor}{RGB}{255, 200, 150}     
\definecolor{thirdcolor}{RGB}{240, 240, 140}      

\usepackage{colortbl}
\usepackage{xcolor}
\usepackage{rotating}
\usepackage{threeparttable}
\usepackage{hyperref}

\newcommand{\best}[1]{\cellcolor[HTML]{AEDCB5}#1}     
\newcommand{\sbest}[1]{\cellcolor[HTML]{FECBA1}#1}    
\newcommand{\tbest}[1]{\cellcolor[HTML]{E2F531}#1}    

\author{Zaid Nasser$^{1*}$, Mikhail Iumanov$^{1*}$, Tianhao Li$^{1*}$, \\ Maxim Popov$^{1}$,  Jaafar Mahmoud$^{1\dagger}$, Sergey Kolyubin$^{1}$
\thanks{$^{*}$ Equal contribution
        }%
\thanks{$^{\dagger}$ Corresponding author: jaafar.a.mahmoud@itmo.ru}
\thanks{$^{1}$ Biomechatronics and Energy-Efficient Robotics (BE2R) Lab, ITMO
University, Saint Petersburg, Russia
        }%
}
\newcommand{\name}{RADIO-ViPE}
\usepackage{threeparttable}  
\usepackage{xcolor}
\usepackage{pifont}

\definecolor{traditional}{RGB}{0, 150, 150}  
\definecolor{semslam}{RGB}{0, 120, 150}  
\definecolor{semantic}{RGB}{230, 159, 0}     
\definecolor{neural}{RGB}{153, 51, 255}      
\definecolor{ours}{RGB}{0, 128, 0}           
\definecolor{darkcyan}{rgb}{0.0, 0.2, 0.13}
\usepackage{graphicx}
\usepackage{booktabs}
\begin{document}


\maketitle
\begin{abstract}
We present \textcolor{magenta}{\name{}} (\textcolor{magenta}{R}educe \textcolor{magenta}{A}ll \textcolor{magenta}{D}omains \textcolor{magenta}{I}nto \textcolor{magenta}{O}ne --- \textcolor{magenta}{Vi}deo \textcolor{magenta}{P}ose \textcolor{magenta}{E}ngine), an online semantic SLAM system that enables geometry-aware open-vocabulary grounding, associating arbitrary natural language queries with localized 3D regions and objects in dynamic environments.
Unlike existing approaches that require calibrated, posed RGB-D input, \name{} operates directly on raw monocular RGB video streams, requiring no prior camera intrinsics, depth sensors, or pose initialization. 
The system tightly couples multi-modal embeddings — spanning vision and language — derived from agglomerative foundation models (e.g., RADIO~\cite{radio}) with geometric scene information. This coupling takes place in initialization, optimization and factor graph connections to improve the consistency of the map from multiple modalities.
The optimization is wrapped within adaptive robust kernels, designed to handle both actively moving objects and agent-displaced scene elements (e.g., furniture rearranged during ego-centric session).
Experiments demonstrate that \name{} achieves state-of-the-art results on the dynamic TUM-RGBD benchmark~\cite{tum-rgbd} while maintaining competitive performance against offline open-vocabulary methods that rely on calibrated data and static scene assumptions. 
\name{} bridges a critical gap in real-world deployment, enabling robust open-vocabulary semantic grounding for autonomous robotics and unconstrained in-the-wild video streams -- \href{https://be2rlab.github.io/radio_vipe}{Project Page}.
\end{abstract}    
\section{Introduction}
\label{sec:intro}

Knowledge mapping — the process of grounding free-form linguistic concepts onto a 3D geometric representation of the environment — is a fundamental capability for general-purpose robots. Unlike humans, who innately perceive objects not merely as geometric primitives but with rich semantic understanding of their identity, function, and affordances, robots have traditionally relied on purely geometric maps for self-localization and collision avoidance. Recent advances in computer vision, foundation models and large language models have begun to bridge this gap, enabling knowledge-enriched scene representations that support object-level reasoning and open-ended task execution. Open-vocabulary semantic perception is a key component here, as it liberates robots from the fixed object taxonomies, allowing flexible, language-driven scene interpretation across diverse, unstructured environments.

A natural source of supervision for such systems is internet-scale video, which offers a much more affordable, scalable, and more diverse alternative to collecting data from curated robot demonstrations, encompassing broad scene layouts, object appearances, and human activities across indoor and outdoor settings~\cite{ego4d, aria}. However, such data is inherently unstructured and uncalibrated, lacking the depth, pose, and semantic annotations required for most of existing 3D scene understanding methods. In addition, real-world environments introduce dynamic disturbances — from moving agents to quasi-static objects such as furniture and appliances displaced by human activity — that corrupt data associations and degrade localization and mapping if left unaddressed.

These challenges collectively motivate \name{}: a calibration-free, online semantic SLAM system that ingests raw, unconstrained monocular RGB video and produces a semantically and geometrically consistent scene representation supporting open-vocabulary language grounding. We build upon ViPE~\cite{vipe} and agglomerative foundation models (\cite{radio, radseg}) that leverage multi-teacher distillation, and provide a unified student model that retains and improves the distinct capabilities of multiple teachers~\cite{radio}. 
\begin{figure}[!t]
    \centering
    \includegraphics[width=0.95\linewidth]{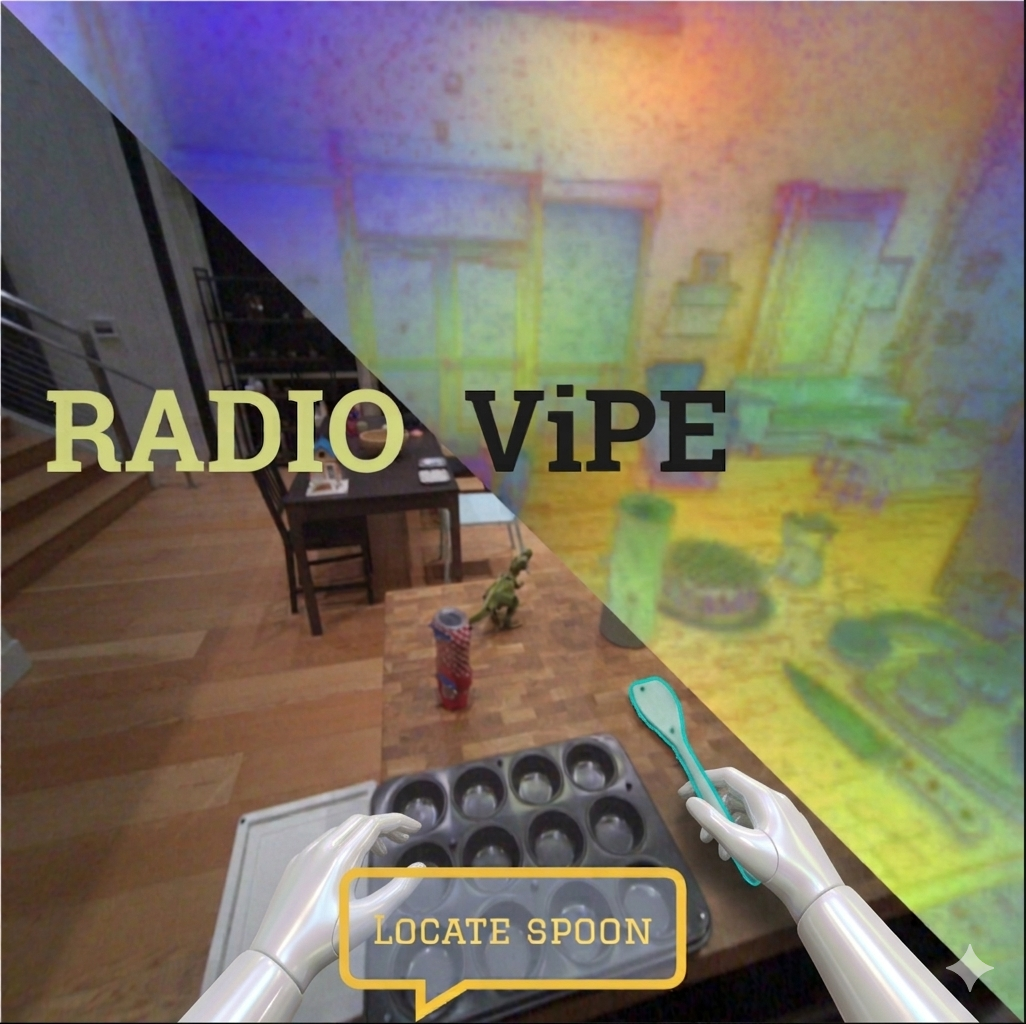}
    \caption{\name{}: An online, ready-to-deploy semantic SLAM system that ground free-form language queries spatially from uncalibrated, unconstrained monocular RGB video}
    \label{fig:teaser}
\end{figure}

The main contributions of this work are stated as follows:
\begin{enumerate}
\item \textbf{Vision-Language-Geometric Fusion via Dense Bundle Adjustment.} A novel tightly coupled multi-modal fusion that jointly embeds high-level features from agglomerative foundation models~\cite{radio} with geometric constraints directly within a dense bundle adjustment framework. The factor graph is also formulated with both geometric-semantic criteria ensuring consistency of the map built from multiple sources of modalities.
\item \textbf{Temporally Consistent Adaptive Robust Kernel.} A dynamic-aware robust optimization scheme that extends adaptive kernel formulations~\cite{rvwo, barron} with temporal consistency, jointly reasoning over geometric reprojection error and cross-view semantic embedding discrepancy to suppress the influence of moving and quasi-static scene elements at data association and improve robustness.
\item \textbf{Calibration-Free Online Open-Vocabulary Semantic SLAM.} An online, ready-to-deploy semantic SLAM system that unifies vision, language, and geometry into a spatial representation, supporting free-form natural language grounding in 3D directly from uncalibrated, unconstrained monocular RGB video — with no depth sensors, pose priors, or category-specific supervision required.
\end{enumerate}

By unifying these concepts, \name{} offers a powerful alternative to popular state-of-the-art feedforward SLAM methods \cite{mast3r, vggt, dust3r}, excelling particularly in dynamic environments.

\section{Related works}
\label{sec:related_works}

SLAM systems have evolved from purely geometric pipelines toward rich, semantically-aware scene understanding. We survey existing approaches along five axes — localization, mapping, semantic understanding, real-time operation, and dynamic scene handling — and organize them into four categories (Table~\ref{tab:comparison}).
\subsection{Geometric SLAM Systems}
Early SLAM systems prioritized geometric accuracy at the expense of semantic reasoning. ORB-SLAM3~\cite{ORB-SLAM3} remains a strong baseline for visual-inertial odometry, offering robust multi-map management and loop closure across diverse sensor configurations, yet provides no semantic interpretation or dynamic object handling. RVWO~\cite{rvwo} addresses dynamic objects in wheeled robot settings but lacks semantic awareness. Kimera~\cite{kimera} and RGBDS-SLAM~\cite{rgbdsslam} incorporate instance-level semantics and RGB-D fusion respectively, yet both remain confined to closed-set, pre-defined object categories. SamSLAM~\cite{SamSLAM} advances dynamic handling via category-agnostic segmentation but cannot associate scene elements with open-vocabulary language descriptions.
\subsection{Offline Open-Vocabulary Scene Understanding}
Foundation models have enabled compelling open-vocabulary understanding in 3D scene reconstruction, albeit exclusively in offline settings. BBQ~\cite{bbq} and ConceptGraphs~\cite{concept-graphs} leverage large vision-language models to construct semantically rich 3D scene graphs with natural language grounding. HOV-SG~\cite{hovsg} further organizes these into hierarchical spatial-semantic structures. OpenScene~\cite{openscene} and OpenMask3D~\cite{openmask3d} achieve zero-shot 3D segmentation by distilling CLIP features into 3D point representations. Despite strong semantic capabilities, all these methods operate offline, lack integrated odometry, and assume static scenes — significantly constraining their applicability in real robotic deployments.
\subsection{Real-Time Open-Vocabulary SLAM}
Recent efforts have sought to reconcile open-vocabulary understanding with real-time operation. CLIO~\cite{clio} introduces an information-theoretic framework that dynamically clusters 3D primitives according to task-driven language instructions. OVO-SLAM~\cite{OVO-SLAM} integrates CLIP embeddings within a Gaussian splatting representation for real-time open-vocabulary mapping. RayFronts~\cite{rayfronts} depends on dense language-aligned features to provide dense voxel-wise embedding of the map. While these systems successfully couple semantic understanding with odometry and mapping, neither provides robustness to dynamic or quasi-static scene disturbances.
\begin{table}[t]
  \centering
  \small
  \begin{threeparttable}

    \begin{tabular}{@{}lccccccc@{}}
      \toprule
      \textbf{Method} & \rotatebox{90}{Online} & \rotatebox{90}{Semantics} & \rotatebox{90}{Grounding} & \rotatebox{90}{Odometry} & \rotatebox{90}{Mapping} & \rotatebox{90}{Dynamic} & \rotatebox{90}{Calib-free}  \\
      \midrule
      
      \textcolor{traditional}{ORBS3}~\cite{ORB-SLAM3}       & \textcolor{olive}{\checkmark} & \textcolor{red}{\ding{55}}    & \textcolor{red}{\ding{55}}    & \textcolor{olive}{\checkmark} & \textcolor{olive}{\checkmark} & \textcolor{red}{\ding{55}} & \textcolor{red}{\ding{55}} \\
      \textcolor{traditional}{RVWO}~\cite{rvwo}       & \textcolor{olive}{\checkmark} & \textcolor{red}{\ding{55}}    & \textcolor{red}{\ding{55}}    & \textcolor{olive}{\checkmark} & \textcolor{olive}{\checkmark} & \textcolor{olive}{\checkmark} &\textcolor{red}{\ding{55}} \\
      \textcolor{traditional}{Kimera}~\cite{kimera}             & \textcolor{olive}{\checkmark} & \textcolor{red}{Closed}       & \textcolor{red}{\ding{55}}    & \textcolor{olive}{\checkmark} & \textcolor{olive}{\checkmark} & \textcolor{red}{\ding{55}} &\textcolor{red}{\ding{55}} \\
      \textcolor{traditional}{SamS}~\cite{SamSLAM}                & \textcolor{olive}{\checkmark} & \textcolor{blue}{Agnostic}     & \textcolor{red}{\ding{55}}    & \textcolor{olive}{\checkmark} & \textcolor{olive}{\checkmark} & \textcolor{olive}{\checkmark} &\textcolor{red}{\ding{55}}\\
      \textcolor{traditional}{RGBDS}~\cite{rgbdsslam}      & \textcolor{olive}{\checkmark} & \textcolor{red}{Closed}       & \textcolor{red}{\ding{55}}    & \textcolor{olive}{\checkmark} & \textcolor{olive}{\checkmark} & \textcolor{red}{\ding{55}} &\textcolor{red}{\ding{55}}\\
      \midrule
      
      \textcolor{semantic}{BBQ}~\cite{bbq}                      & \textcolor{red}{\ding{55}}      & \textcolor{green}{Open}     & \textcolor{olive}{\checkmark} & \textcolor{red}{\ding{55}}      & \textcolor{olive}{\checkmark} & \textcolor{red}{\ding{55}}&\textcolor{red}{\ding{55}} \\
      \textcolor{semantic}{CG}~\cite{concept-graphs} & \textcolor{red}{\ding{55}}      & \textcolor{green}{Open}     & \textcolor{olive}{\checkmark} & \textcolor{red}{\ding{55}}      & \textcolor{olive}{\checkmark} & \textcolor{red}{\ding{55}} &\textcolor{red}{\ding{55}}\\
      \textcolor{semantic}{HOV-SG}~\cite{hovsg}        & \textcolor{red}{\ding{55}}      & \textcolor{green}{Open}     & \textcolor{olive}{\checkmark} & \textcolor{red}{\ding{55}}      & \textcolor{olive}{\checkmark} & \textcolor{red}{\ding{55}} &\textcolor{red}{\ding{55}}\\
      
      \textcolor{semantic}{OS}~\cite{openscene}            & \textcolor{red}{\ding{55}}      & \textcolor{green}{Open}     & \textcolor{olive}{\checkmark} & \textcolor{red}{\ding{55}}      & \textcolor{red}{\ding{55}}    & \textcolor{red}{\ding{55}} &\textcolor{red}{\ding{55}}\\
      \textcolor{semantic}{OM3D}~\cite{openmask3d}          & \textcolor{red}{\ding{55}}      & \textcolor{green}{Open}     & \textcolor{olive}{\checkmark} & \textcolor{red}{\ding{55}}      & \textcolor{red}{\ding{55}}    & \textcolor{red}{\ding{55}} &\textcolor{red}{\ding{55}}\\
      \midrule

      \textcolor{neural}{CLIO}~\cite{clio}                    & \textcolor{olive}{\checkmark} & \textcolor{orange}{Task-driven} & \textcolor{olive}{\checkmark} & \textcolor{olive}{\checkmark} & \textcolor{olive}{\checkmark} & \textcolor{red}{\ding{55}} &\textcolor{red}{\ding{55}}\\
      \textcolor{neural}{OVO}~\cite{OVO-SLAM}         & \textcolor{olive}{\checkmark} & \textcolor{green}{Open}     & \textcolor{olive}{\checkmark} & \textcolor{olive}{\checkmark} & \textcolor{olive}{\checkmark} & \textcolor{red}{\ding{55}} &\textcolor{red}{\ding{55}}\\
      \textcolor{neural}{Rayfronts}~\cite{rayfronts}                    & \textcolor{olive}{\checkmark} & \textcolor{green}{Open} & \textcolor{olive}{\checkmark} & \textcolor{red}{\ding{55}} & \textcolor{olive}{\checkmark} & \textcolor{red}{\ding{55}} &\textcolor{red}{\ding{55}}\\

      \midrule
      \textcolor{purple}{VGGT}~\cite{vggt}& 
      \textcolor{olive}{\checkmark} & \textcolor{red}{\ding{55}}      & \textcolor{red}{\ding{55}}  & \textcolor{olive}{\checkmark} & \textcolor{olive}{\checkmark} & \textcolor{red}{\ding{55}}  &\textcolor{olive}{\checkmark}\\
      \textcolor{purple}{DUST3R}~\cite{dust3r}                          & 
      \textcolor{olive}{\checkmark} & \textcolor{red}{\ding{55}}      & \textcolor{red}{\ding{55}}  & \textcolor{olive}{\checkmark} & \textcolor{olive}{\checkmark} & \textcolor{olive}{\checkmark} &\textcolor{olive}{\checkmark}\\
      \textcolor{purple}{ViPE}~\cite{vipe}                          & 
      \textcolor{olive}{\checkmark} & \textcolor{orange}{Predefined}      & \textcolor{red}{\ding{55}}  & \textcolor{olive}{\checkmark} & \textcolor{olive}{\checkmark} & \textcolor{red}{\ding{55}} &\textcolor{olive}{\checkmark}\\
      \midrule
      \textcolor{magenta}{\textbf{\name{}}}                           & 
      \textcolor{olive}{\checkmark} & \textcolor{green}{Open}     & \textcolor{olive}{\checkmark} & \textcolor{olive}{\checkmark} & \textcolor{olive}{\checkmark} & \textcolor{olive}{\checkmark} &\textcolor{olive}{\checkmark}\\
      \bottomrule
    \end{tabular}
    \caption{Comparison of State-of-the-Art SLAM Methods}
    \label{tab:comparison}
    \begin{tablenotes}
    \item \textbf{Columns} indicate: Online operation, Semantic understanding type (Closed=predefined classes, Agnostic=category-agnostic without semantics, Open=open-vocabulary), Geometric Grounding=the ability to textually access and interact with the 3$\mathrm{D}$ map, Odometry estimation, Map reconstruction, Dynamic scene handling and calibration requirement.
    \end{tablenotes}
  \end{threeparttable}
\end{table}
\subsection{Feed-Forward SLAM}
Feed-forward approaches directly regress 3D geometry from images. DUSt3R~\cite{dust3r} reformulates pairwise reconstruction as pointmap regression, removing rigid projective camera assumptions. VGGT-SLAM~\cite{vggt} incrementally aligns submaps while explicitly optimizing over the SL(4) manifold to resolve the 15-DoF projective ambiguity inherent to uncalibrated monocular reconstruction. ViPE~\cite{vipe} extends this paradigm to in-the-wild video, achieving robust intrinsic and extrinsic estimation with near-metric depth at 3--5 FPS. However, feed-forward systems broadly assume scene rigidity, making them vulnerable to dynamic objects, and lack the high-level semantic representations required for open-vocabulary grounding.

Despite substantial progress across these paradigms, no existing system simultaneously achieves real-time operation, open-vocabulary semantic grounding, robust odometry, accurate mapping, and dynamic scene robustness. This integration gap severely limits deployment in complex, real-world environments. \name{} addresses this gap with a unified framework that seamlessly combines all these capabilities — supporting applications ranging from autonomous robot navigation and ego-centric~\cite{aria} or scene-centric~\cite{replica} video data resourcing.

\section{Methodology}
\begin{figure*}[!h]
    \centering
    \includegraphics[width=0.85\textwidth]{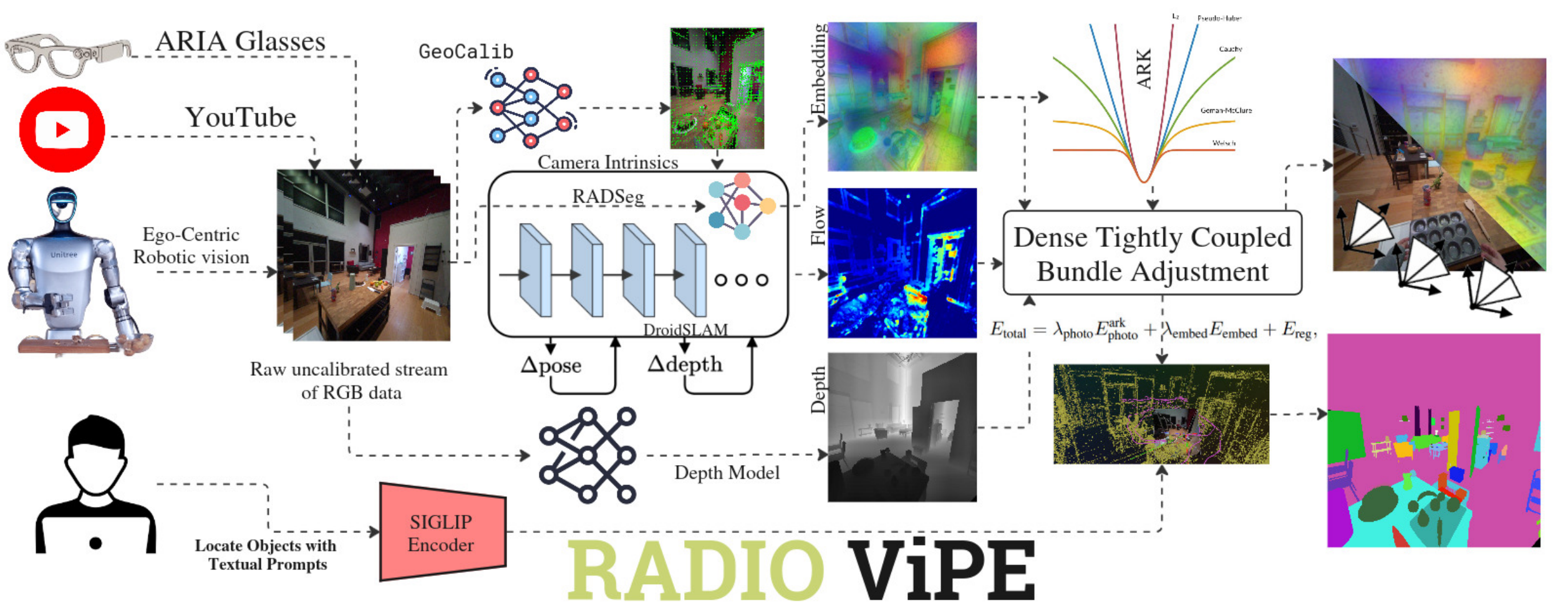} 
    \caption{\name{} pipeline}
    \label{fig:pipeline}
\end{figure*}

\name{} accepts an uncalibrated monocular RGB video stream and unifies camera pose optimization from flow estimation based on ~\cite{droid}, depth from monocular foundation depth models~\cite{unidepth, moge} that is aligned with consistent 3D SLAM point cloud, and dense high-level visual embeddings from RADIO~\cite{radio} within a sliding window factor graph optimization framework. The optimization is then wrapped with adaptive kernels~\cite{rvwo,ark} to improve robustness in dynamic environments. Altogether the pipeline is operating online at approximately 8--10 FPS.
\subsection{System Overview}

The pipeline is based on \cite{vipe,droid} but with the following adjustments as follows (See Fig.~\ref{fig:pipeline}):

\begin{enumerate}
    \item \textbf{Camera Initialization.} Camera intrinsics are bootstrapped from uniformly sampled frames using GeoCalib~\cite{veicht2024geocalib}, requiring no calibration targets or known camera models, and are subsequently co-optimized within bundle adjustment~\cite{vipe}.
    
    \item \textbf{Keyframe Selection.} Relative motion between each incoming frame and the most recent keyframe is estimated via weighted dense optical flow~\cite{droid, vipe}. Frames exceeding a predefined motion threshold are designated as keyframes and appended to the factor graph $\mathcal{G} = (\mathcal{V}, \mathcal{E})$, where $\mathcal{V}$ denotes keyframes and $\mathcal{E}$ pairwise connections.

    \item \textbf{Multi-Modal Feature Extraction.} Dense multi-modal embeddings are extracted per keyframe using RADSeg~\cite{radio, radseg} (Sec.~\ref{sec:radseg}), upsampled via bilinear interpolation to $(H/8, W/8)$, and compressed to $D{=}256$ dimensions via PCA (Sec.~\ref{sec:experiments}) to ensure scalability and efficiency in memory management.

    \item \textbf{Depth Estimation.} A metric depth map is estimated per keyframe using monocular foundation depth models~\cite{moge, unidepth, metric3dv2}, converted to inverse depth (disparity) for numerical stability, and downsampled by a factor of 8 to match the estimated optical flow resolution~\cite{vipe}.

\item \textbf{Semantic Flow Initialization.} To improve the robustness of optical flow priors ~\cite{droid} in textureless surfaces, we augment the geometric reprojection prior with a semantic correspondence term derived from dense RADIO features. Concretely, for each pixel $\mathbf{u}$ in frame $i$, we compute cosine similarities between the PCA-compressed Radio embedding $\mathbf{Z}_i(\mathbf{u}) \in \mathbb{R}^K$ and pixels embeddings $\mathbf{Z}_j(\mathbf{v})$, yielding a dense semantic flow field $\boldsymbol{\Omega}^{\text{sem}}(\mathbf{u})$. The two priors are then fused via per-pixel confidence-based blending:
\begin{equation}
    \boldsymbol{\Omega}^{\text{prior}}(\mathbf{u})
    \;:=\;
    \beta\,\boldsymbol{\Omega}^{\text{prior}}(\mathbf{u})
    \;+\;
    \bigl(1 - \beta\bigr)\,\boldsymbol{\Omega}^{\text{sem}}(\mathbf{u}),
    \label{eq:blended_prior}
\end{equation}
where the blending weight $\beta$ balances the photometric flow confidence produced by the flow network~\cite{droid} against the peak semantic similarity score. The blended initialization replaces $\boldsymbol{\Omega}^{\text{prior}}$ when constructing the correlation volume for the photometric flow term, leaving the flow network architecture unchanged.
    \item \textbf{Bundle Adjustment.} Camera intrinsics, poses, and 3D scene structure are jointly refined over all keyframes by minimizing a vision-language-geometric energy function (Sec.~\ref{sec:ba}). All terms are evaluated on downsampled images $\tilde{\mathbf{I}} \in \mathbb{R}^{H \times W \times 3}$, $H {=} h/8$, $W {=} w/8$, to ensure compatibility with the estimated optical flow~\cite{vipe}. Factor graph connectivity is augmented beyond geometric proximity by \emph{embedding-based co-visibility}: a global descriptor per keyframe is obtained by mean-pooling its RADSeg embeddings and $\ell_2$-normalizing the result, yielding a compact representation that is free to compute as the embeddings are already resident in the keyframe buffer. Incoming keyframes are matched against all non-recent keyframes (excluding the most recent $\tau$ frames to suppress trivially adjacent matches) via a single cosine-similarity query; pairs whose similarity exceeds a threshold $\eta$ are linked by bidirectional edges injected into the factor graph.

    \item \textbf{Non-Keyframe Pose Estimation.} Each non-keyframe is connected to its two nearest keyframes via uni-directional edges, and its pose is recovered through photometric alignment, bypassing per-frame depth estimation. All non-keyframe poses are estimated in parallel~\cite{vipe}.

    \item \textbf{Open-Vocabulary Grounding.}
    Real-time Open-Vocabulary capabilities are achieved via decoding the compressed Radio features of 3D points and then projecting them into SigLip~\cite{siglip} latent space for matching with the text query embedding

\end{enumerate}

\subsection{Dense Visual Features Extraction}
\label{sec:radseg}
We present a dense spatial embedding pipeline that integrates and adapts existing components to achieve robust feature representation tailored for dynamic environments. Our approach employs RADSeg~\cite{radseg} as the core dense embedding model, while we use its agglomerative capabilities—grounded in RADIO~\cite{radio}—to generate language-aligned features within the SigLIP~\cite{siglip} embedding space. 

Furthermore, we operate RADSeg using a sliding-window approach, performing inference over overlapping image regions and subsequently refining the aggregated feature map through a self-attention mechanism. It allowed us to provide an optimal balance between spatial discriminability and semantic language alignment. 

Unlike conventional approaches that operate on the language-aligned space, we perform PCA directly on the encoder feature space, thereby preserving structural integrity that is essential for spatial reasoning. To ensure a representative PCA mapping, we exploit the initialization period of our Bundle Adjustment (BA) optimization. By waiting until a sufficient buffer of keyframes is collected before computing the PCA components, we achieve a robust representation that balances computational efficiency with feature expressivity.

\subsection{Joint Bundle Adjustment}
\label{sec:ba}

Let $\mathbf{T}_i \in \text{SE}(3)$ denote the world-to-camera pose for frame $i$, and $\mathbf{K}_q = [f_x, f_y, c_x, c_y]^\top$ the camera intrinsics. \name{} formulates dense SLAM as a factor graph optimization, jointly refining poses $\{\mathbf{T}_i\}$, dense disparity maps $\{\mathbf{d}_i\}$, and intrinsics $\{\mathbf{K}_q\}$ by minimizing a weighted vision-language-geometric energy function. 

\subsubsection{Dense Photometric Flow Term}

Following DROID-SLAM~\cite{droid}, geometric consistency is enforced via dense optical flow constraints. For each edge $(i,j)$ in $\mathcal{G}$, pixels $\mathbf{u}$ in frame $i$ are projected into frame $j$ as:
\begin{equation}
    \mu_{ij} = \Pi_j\!\left(\mathbf{T}_j \mathbf{T}_i^{-1} \circ \Pi_i^{-1}(\mathbf{u},\, d_i(\mathbf{u}))\right),
\end{equation}
where $\Pi_q$ and $\Pi_q^{-1}$ denote the projection and unprojection functions under intrinsics $\mathbf{K}_q$. An optical flow network~\cite{droid} predicts a residual dense flow field $\boldsymbol{\Omega}_{ij} \in \mathbb{R}^{H \times W \times 2}$ alongside per-pixel confidence weights $w(\mathbf{u})$ conditioned on the relative poses of the edge's frames. The prior flow estimate $\boldsymbol{\Omega}_{ij}^{\text{prior}} = \mu_{ij} - \mathbf{u}$ initializes a correlation volume for iterative refinement. The photometric term is then:
\begin{equation}
\label{eq:photo_term}
    E_{\text{photo}} = \sum_{\mathbf{u}} w(\mathbf{u}) \cdot \left\| \boldsymbol{\Omega}_{ij}^{\text{prior}} - \boldsymbol{\Omega}_{ij}(\mathbf{u}) \right\|^2,
\end{equation}
which directly penalizes discrepancies between geometric reprojections and learned dense correspondences.

\subsubsection{RADIO Embedding Similarity Term}

To incorporate semantic consistency into the optimization, we introduce a novel embedding similarity term that enforces cross-view feature alignment under geometric constraints, enabling \name{} to leverage the rich multi-modal representations of RADIO~\cite{radio, radseg} directly within bundle adjustment.

For each edge $(i,j)$, let $\mathbf{Z}_i, \mathbf{Z}_j \in \mathbb{R}^{K \times H \times W}$ denote the PCA-compressed dense embedding maps. Each source pixel $\mathbf{u} \in I_i$ is projected into the target frame using current pose and depth estimates to locate its corresponding pixel $\mathbf{v} =P_{i,j}(\mathbf{u})\in I_j $, then the target embedding is recovered via bilinear interpolation:
\begin{equation}
    \hat{\mathbf{Z}}_j(P_{i,j}(\mathbf{u})) = \sum_{(l,m)\,\in\,\mathcal{N}(P_{i,j}(\mathbf{u}))} w_{lm}(P_{i,j}(\mathbf{u})) \cdot \mathbf{Z}^{j}_{lm}
\end{equation}
where $\mathcal{N}(P_{i,j}(\mathbf{u}))$ denotes the four nearest integer neighbours of $P_{i,j}(\mathbf{u}) = (u,v)$, with bilinear weights:
\begin{align*}
    w_{lm}(P_{i,j}(\mathbf{u})) &\in \bigl\{\alpha\beta,\ \alpha(1{-}\beta),\ (1{-}\alpha)\beta,\ (1{-}\alpha)(1{-}\beta)\bigr\}, \\
    \quad \alpha &= u - \lfloor u \rfloor,\ \ \beta = v - \lfloor v \rfloor
\end{align*}

After $\ell_2$-normalizing both embeddings, the cosine similarity between the source and projected target is:
\begin{equation}
    cs_{ij}(\mathbf{u}) = \frac{\mathbf{Z}_i(\mathbf{u})^\top\, \hat{\mathbf{Z}}_j(P_{i,j}(\mathbf{u}))}{\|\mathbf{Z}_i(\mathbf{u})\| \cdot \|\hat{\mathbf{Z}}_j(P_{i,j}(\mathbf{u}))\|},
\end{equation}

The embedding residual is casted in photometric form to maintain consistency with the flow term:
\begin{equation}
    r_{\text{embed}}(\mathbf{u}) = \lambda_{\text{embed}}\,\sqrt{2\bigl(1 - cs_{ij}(\mathbf{u})\bigr)},
\end{equation}
with $\lambda_{\text{embed}} {=} 2$, yielding residuals of comparable magnitude to intensity differences in classical photometric SLAM. The full embedding similarity term is:
\begin{equation}
    E_{\text{embed}} = \sum_{\mathbf{u}} w(\mathbf{u}) \cdot r_{\text{embed}}^2(\mathbf{u})
\end{equation}
where $w(\mathbf{u})$ incorporates photometric flow confidence.




\subsection{Temporally Consistent Adaptive Robust Kernel}
\label{sec:ark}
One of the core contribution of our method is introducing adaptive robust kernels to improve SLAM robustness in dynamic environments. Since a pairwise kernel shaped solely by the cosine
similarity $cs_{ij}(\mathbf{u})$ across a \emph{single} edge $(i,j)$ cannot distinguish genuinely static surfaces, movable objects, and actively moving agents — reliable classification requires aggregating evidence \emph{across multiple edges temporally}.

\subsubsection{Temporal Stability Field}
Let $\mathcal{N}(i)$ denote all keyframes connected to keyframe $i$ in
$\mathcal{G}$. Using the per-edge cosine similarities $cs_{ij}(\mathbf{u})$
defined in Sec.~\ref{sec:ba}-2, we compute the temporal mean and variance at
each pixel:
\begin{align*}
    \bar{cs}_i(\mathbf{u}) &=
        \frac{1}{|\mathcal{N}(i)|}
        \sum_{j \in \mathcal{N}(i)} cs_{ij}(\mathbf{u}), \\
    \sigma^2_i(\mathbf{u}) &=
        \frac{1}{|\mathcal{N}(i)|}
        \sum_{j \in \mathcal{N}(i)}
        \bigl(cs_{ij}(\mathbf{u}) - \bar{cs}_i(\mathbf{u})\bigr)^2
\end{align*}
The \emph{temporal stability field} is then defined as:
\begin{equation}
    \mathcal{S}_i(\mathbf{u}) =
        \bar{cs}_i(\mathbf{u})
        \cdot
        \bigl(1 - \sigma^2_i(\mathbf{u})\bigr)
        \;\in\; [0,\,1]
    \label{eq:stability_field}
\end{equation}
$\mathcal{S}_i(\mathbf{u}) \approx 1$ indicates a genuinely static surface
--- consistently high similarity with low temporal variance.
$\mathcal{S}_i(\mathbf{u}) \approx 0$ flags either active motion (low mean)
or a displaced object (high variance), correctly identifying both as
unreliable. 

\subsubsection{Three-Regime Barron Shape Mapping}
\begin{figure}[t]
    \centering
    \includegraphics[width=0.45\textwidth]{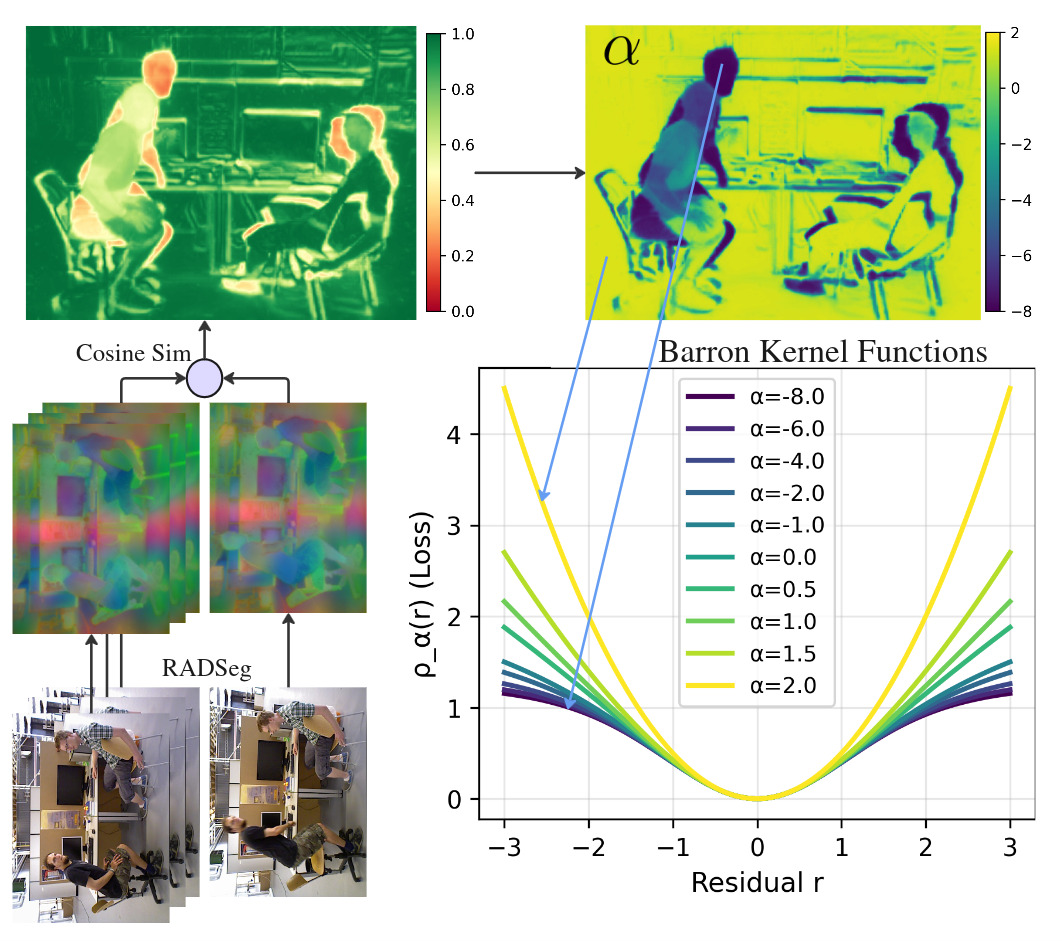}
    \caption{
    Adaptive robust kernels based on Barron's general loss~\cite{barron}.
             The shape parameter $\alpha$ is governed by the temporal stability
             field $\mathcal{S}_i(\mathbf{u})$, transitioning from $\ell_2$
             (static surfaces) through Huber (movable objects) to Cauchy
             (actively moving agents).
             }
    \label{fig:barron}
\end{figure}

We classify pixels in real-world video to 3 physically distinct categories
that require different loss regimes: genuinely static, movable objects by agent (e.g. in ego-centric session), and actively moving agents
(e.g. people). So, we map $\mathcal{S}_i(\mathbf{u})$ to the shape parameter $\alpha$ of the
general Barron loss $\rho_\alpha$~\cite{barron}, which recovers $\ell_2$ ($\alpha{=}2$), Huber ($\alpha{=}1$), and Cauchy
($\alpha{\to}0$) as special cases. We assign $\alpha$ via a differentiable
piecewise-linear mapping:
\begin{equation}
    \alpha_i(\mathbf{u}) =
    \begin{cases}
        2,
            & \mathcal{S}_i(\mathbf{u}) \geq \theta_s \\[4pt]
        1
            + \dfrac{\mathcal{S}_i(\mathbf{u}) - \theta_m}{\theta_s - \theta_m}
              ,
            & \theta_m \leq \mathcal{S}_i(\mathbf{u}) < \theta_s \\[8pt]
        \alpha_{\mathrm{dyn}}
            + \dfrac{\mathcal{S}_i(\mathbf{u})}{\theta_m}
              \bigl(1 - \alpha_{\mathrm{dyn}}\bigr),
            & \mathcal{S}_i(\mathbf{u}) < \theta_m
    \end{cases}
    \label{eq:three_regime}
\end{equation}
with $\alpha_{\mathrm{dyn}}{\leq}0$, $\theta_s{=}0.75$, and $\theta_m{=}0.35$.
The three regimes apply the correct loss family to each category, while keeping the shape parameter $\alpha$ continuous and differentiable.
We then rewrite \eqref{eq:photo_term} as follows:
\begin{align}
    E^{\mathrm{ark}}_{\mathrm{photo}} &=
    \sum_{\mathbf{u}}
    w_{ark}
    \!\left(E_{\mathrm{photo}}(\mathbf{u}),\alpha_{i}\right)\cdot E_{\mathrm{photo}}(\mathbf{u}), \\
    w_{\mathrm{ark}}(r,\alpha) &=
        \frac{1}{\max(r,\varepsilon)}
        \frac{\partial\,\rho_\alpha(r)}{\partial r},
\end{align}
where $\varepsilon$ prevents division by zero, and $\rho_\alpha$ is the barron function~\cite{barron}(Fig.~\ref{fig:barron}). The adaptive robust kernel thus jointly down-weights low-confidence and semantically dynamic regions via $w$ and $\alpha_{ij}(\mathbf{u})$ respectively, while preserving full $\ell_2$ influence on static scene structure (See Fig.~\ref{fig:barron}).

\subsection{Objective for Tightly-Coupled Optimization}

Finally, we perform optimization with Gauss--Newton method to jointly calculate camera poses $\{\mathbf{T}_i\}$, disparities $\{\mathbf{d}_i\}$, and intrinsics $\{\mathbf{K}_q\}$ by minimizing the following complete objective function:

\begin{align}
E_{\text{total}} &=
\gamma_{\text{photo}} E_{\text{photo}}^{\text{arc}}
+ \gamma_{\text{embed}} E_{\text{embed}}
+ E_{\text{reg}} \quad |\\
E_{\text{reg}}(\mathbf{d}_i) &= \alpha_{\text{disp}} \sum_{\mathbf{u}} \| \mathbf{d}_i(\mathbf{u}) - \mathbf{d}_i^{\text{prior}}(\mathbf{u}) \|^2,
\end{align}
where $E_{\text{reg}}$ is the regularization term with disparity $\mathbf{d}_i^{\text{prior}}$ from the foundation depth model and the regularization weight $\alpha_{\text{disp}}$ (for experiments we set it to 1.0) that is introduced to stabilize depth estimates and incorporate prior knowledge from foundation depth models similarly to ~\cite{vipe}. This term prevents drift while allowing the optimization to refine depth estimates based on multi-view consistency.
\begin{table}[htbp]
    \small
    \setlength{\tabcolsep}{1pt}
    \caption{SLAM Performance Comparison on TUM-RGBD in cm (ATE $\downarrow$)}
    \label{tab:slam-tum}
    \centering
    \begin{threeparttable}
        \begin{tabular}{lcccccccc|c}
            \toprule
            \textbf{Method}
            & \rotatebox{90}{\textbf{fr3/w/xyz}}
            & \rotatebox{90}{\textbf{fr3/w/rpy}}
            & \rotatebox{90}{\textbf{fr3/w/hs}}
            & \rotatebox{90}{\textbf{fr3/w/static}}
            & \rotatebox{90}{\textbf{fr3/s/xyz}}
            & \rotatebox{90}{\textbf{fr3/s/rpy}}
            & \rotatebox{90}{\textbf{fr3/s/hs}}
            & \rotatebox{90}{\textbf{fr3/s/static}}
            & \rotatebox{90}{\textbf{Average}} \\
            \midrule
            Dyna-SLAM~\cite{dynaslam2}  & 1.64 & 3.54 & 2.96 & 0.68 & 1.27 & --   & 1.86 & --   & 2.00 \\
            DLD-SLAM~\cite{dldslam}     & 1.85 & 4.24 & \tbest{2.19} & 0.56 & --   & --   & --   & --   & 2.21 \\
            V3D-SLAM~\cite{v3d}         & \sbest{1.53} & 7.81 & 2.29 & 0.65 & \best{0.87} & \best{1.69} & \sbest{1.47} & 0.58 & 2.10 \\
            DGS-SLAM~\cite{dgs}         & 4.10 & --   & 5.50 & 0.60 & --   & --   & 4.40 & --   & 3.65 \\
            RoDyn-SLAM~\cite{rodyn}     & 8.30 & --   & 5.60 & 1.70 & --   & --   & 2.70 & --   & 4.58 \\
            DynaMON~\cite{dynamon}    & \best{1.4} & 3.9   & \sbest{2.0} & 1.4 & \sbest{0.9}   & \sbest{2.1}   & 1.9 & \best{0.5}  & \sbest{1.76} \\
            ViPE (SAM)~\cite{vipe}      & 2.4 & \sbest{3.46} & 2.52 & \sbest{0.54} & 1.43 & 3.80 & 2.57 & 0.6 & 2.17 \\
            \text{\name{}}              & 1.90 & \tbest{3.50} & 3.10 & \tbest{0.55} & 1.15 & 2.72 & \tbest{1.60} & \sbest{0.53} & \tbest{1.90} \\
            $\text{\name{}}_{ark}$      & \tbest{1.55} & \best{3.39} & \best{1.96} & \best{0.50} & \tbest{0.98} & \tbest{2.65} & \best{1.44} & \tbest{0.56} & \best{1.63} \\
            \bottomrule
        \end{tabular}
        \begin{tablenotes}
            \footnotesize
            \item \colorbox[HTML]{AEDCB5}{\phantom{xx}} Best \quad
                  \colorbox[HTML]{FECBA1}{\phantom{xx}} Second best 
                  \colorbox[HTML]{E2F531}{\phantom{xx}} Third best 
                  \quad
                  {--} Not reported
        \end{tablenotes}
    \end{threeparttable}
\end{table}

\begin{table*}[!h]
    \small
    \setlength{\tabcolsep}{5pt}
    \caption{Quantitative comparison onReplica ~\cite{replica} following ~\cite{rayfronts}. The rightmost 4 columns indicate whether a method supports online inference and operates without calibration, depth, or pose inputs.}
    \label{tab:replica}
    \centering
    \begin{threeparttable}
        \begin{tabular}{lcccccccccc}
            \toprule
            &
            \multicolumn{3}{c}{\textbf{Without Background}} 
            & \multicolumn{3}{c}{\textbf{With Background}} 
            & & \textbf{Calib} & \textbf{Depth} & \textbf{Pose} \\
            \textbf{Methods}
            & \textbf{mIoU$\uparrow$} & \textbf{f-mIoU$\uparrow$} & \textbf{Acc$\uparrow$}
            & \textbf{mIoU$\uparrow$} & \textbf{f-mIoU$\uparrow$} & \textbf{Acc$\uparrow$}
            & \textbf{Online}
            & \textbf{Free}
            & \textbf{Free}
            & \textbf{Free} \\
            \midrule
            \textbf{ConceptFusion}~\cite{concept-fusion}
              & 21.07 & 31.51 & 35.65 & 20.38 & 35.75 & 41.58
              & \textcolor{red}{\ding{55}} & \textcolor{red}{\ding{55}} & \textcolor{red}{\ding{55}} & \textcolor{red}{\ding{55}} \\
            \textbf{ConceptGraphs}~\cite{concept-graphs}
              & 11.63 & 16.61 & 19.80 & 11.72 & 21.35 & 28.28
              & \textcolor{red}{\ding{55}} & \textcolor{red}{\ding{55}} & \textcolor{red}{\ding{55}} & \textcolor{red}{\ding{55}} \\
            \textbf{HOV-SG}~\cite{hovsg}
              & 16.93 & 31.45 & 34.74 & 19.29 & 30.64 & 35.17
              & \textcolor{red}{\ding{55}} & \textcolor{red}{\ding{55}} & \textcolor{red}{\ding{55}} & \textcolor{red}{\ding{55}} \\
            \textbf{NACLIP-3D}~\cite{naclip}
              & 20.37 & 35.08 & 47.47 & 15.30 & 16.98 & 26.23
              & \textcolor{red}{\ding{55}} & \textcolor{red}{\ding{55}} & \textcolor{red}{\ding{55}} & \textcolor{red}{\ding{55}} \\
            \textbf{Trident-3D}~\cite{trident}
              & 21.30 & 43.34 & 54.79 & \tbest{20.63} & \tbest{38.53} & \tbest{50.31}
              & \textcolor{red}{\ding{55}} & \textcolor{red}{\ding{55}} & \textcolor{red}{\ding{55}} & \textcolor{red}{\ding{55}} \\
            \textbf{RayFronts}~\cite{rayfronts}
              & \best{39.37} & \best{62.03} & \best{68.80} & \sbest{27.73} & \sbest{43.37} & \sbest{54.45}
              & \textcolor{olive}{\checkmark} & \textcolor{red}{\ding{55}} & \textcolor{red}{\ding{55}} & \textcolor{red}{\ding{55}} \\
            \bottomrule
            $\text{\textbf{\name{}}}_{GT}$
              & \sbest{29.51} & \sbest{52.24} & \sbest{59.80} & \best{28.19} & \best{54.44} & \best{65.21}
              & \textcolor{olive}{\checkmark} & \textcolor{red}{\ding{55}} & \textcolor{red}{\ding{55}} & \textcolor{red}{\ding{55}} \\
              $\text{\textbf{\name{}}}$
              & \tbest{24.25} & \tbest{50.63} & \tbest{59.25} & 19.00 & 37.13 & 48.38
              & \textcolor{olive}{\checkmark} & \textcolor{olive}{\checkmark} & \textcolor{olive}{\checkmark} & \textcolor{olive}{\checkmark} \\
            \bottomrule
        \end{tabular}
        \begin{tablenotes}
            \footnotesize
            \item \colorbox[HTML]{AEDCB5}{\phantom{xx}} Best \quad
                  \colorbox[HTML]{FECBA1}{\phantom{xx}} Second best
                  \colorbox[HTML]{E2F531}{\phantom{xx}} Third best 
        \end{tablenotes}
    \end{threeparttable}
\end{table*}

\section{Experiments}
\label{sec:experiments}



We evaluate \name{} on open-vocabulary semantic segmentation on Replica~\cite{replica} and robustness in dynamic environments on TUM-RGBD~\cite{tum-rgbd}. We denote the variant using only the embedding error term as $\text{\name{}}$, and the full pipeline with adaptive kernel tuning as $\text{\name{}}_{ark}$. All experiments are conducted on Intel Xeon Gold $5320$ CPU $2.20$ GHz with NVIDIA GeForce RTX $4090$.
\subsection{SLAM Performance}
Our embedding error term and adaptive robust kernel improve SLAM in dynamic environments by compensating for or attenuating incorrect data associations caused by object motion. Experiments on the dynamic TUM RGB-D~\cite{tum-rgbd} sequences show best average ATE, outperforming existing dynamic SLAM methods. Unlike ViPE~\cite{vipe}, which relies on foundation models (Grounding DINO~\cite{gdino}, SAM~\cite{sam}) and manual class specification for dynamic masking, our pipeline achieves better performance with substantially fewer resources.


\subsection{3D Semantic Segmentation}
\begin{figure}
    \centering
    \includegraphics[width=0.45\textwidth]{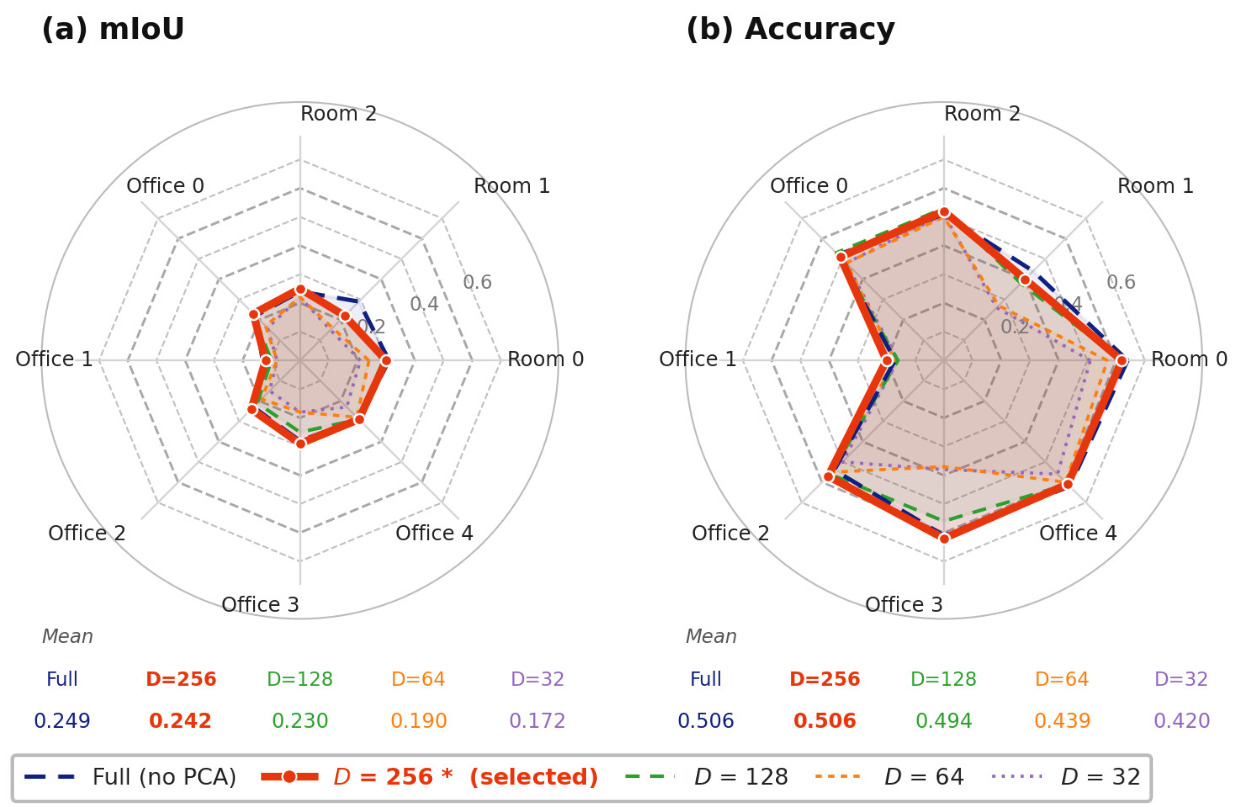}
    \caption{
        \textbf{Ablation study on RADIO PCA feature dimensionality for semantic mapping on the Replica dataset.}
        $D{=}256$ ({\color{red}\textbf{---$\bullet$---}}) closely matches the full-dimensional baseline
        ($\Delta$mIoU $< 1\%$).
        }
    \label{fig:ablation_pca_dim}
\end{figure}

A core feature of our approach is the ability to construct a semantically-aware map during real-time operation. To assess this capability, we evaluate our semantic segmentation performance on the Replica dataset~\cite{replica}. We adopt $D{=}256$ as the default PCA encoded dimension of RADIO features, offering an effective trade-off between memory efficiency and segmentation performance (Fig.~\ref{fig:ablation_pca_dim}).
Unlike RayFronts~\cite{rayfronts}, ConceptFusion~\cite{concept-fusion}, ConceptGraphs~\cite{concept-graphs}, HOV-SG~\cite{hovsg}, NACLIP-3D~\cite{naclip}, and Trident-3D~\cite{trident}, \name{} operates without known camera parameters,
poses, point cloud, or depth. Furthermore, except for RayFronts~\cite{rayfronts},
none of these works support real-time operation;

We follow the evaluation protocol of RayFronts~\cite{rayfronts} on eight Replica
scenes (\texttt{office0--4}, \texttt{room0--2}); results are reported in
Table~\ref{tab:replica}.
Our method ranks in the top-3 among all compared approaches.
The key observation is the small gap between RADIO-ViPE and
RADIO-ViPE$_\text{GT}$: removing ground-truth depth, pose, and calibration
costs only ${\sim}1$--$2\%$ in f-mIoU and Acc (\textbf{without-background} setting),
confirming that our method retains most of its accuracy without \emph{any}
geometric supervision. Performance degrades more noticeably in the
\textbf{with-background} setting, which we attribute to the difficulty of
segmenting structural classes (\emph{e.g.}, wall, floor); we consider this
a current limitation.

\begin{figure}
    \centering
    \includegraphics[width=0.48\textwidth]{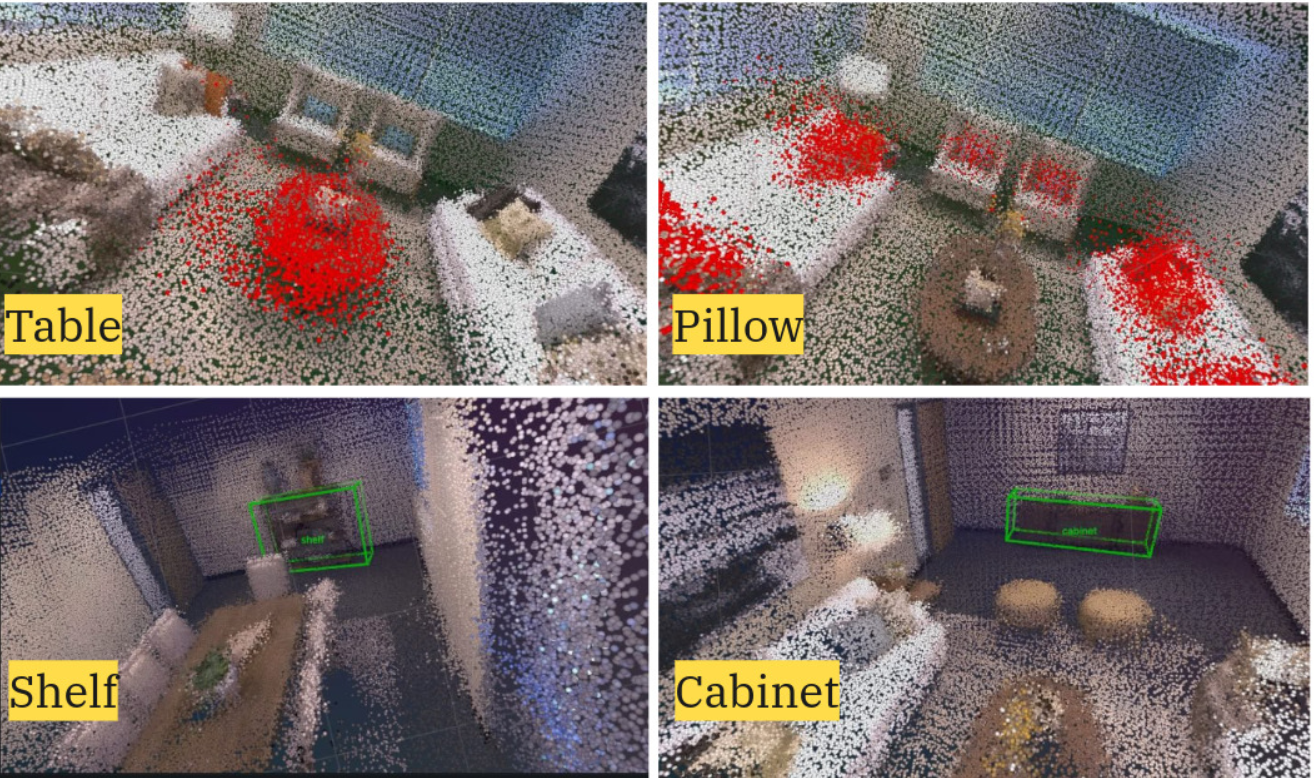} 
    \caption{Quantitative results of \name{} on Replica with different text queries.}
    \label{fig:viz}
\end{figure}
\section{Conclusion}
RADIO-ViPE is a robust, online semantic SLAM system that operates without calibration
and handles dynamic scenes. Leveraging RADSeg features, it achieves competitive 
performance on TUM-RGBD~\cite{tum-rgbd} and ranks top-3 on Replica despite PCA-based memory compression---demonstrating a strong performance-to-efficiency trade-off suitable for real-time robotics, AR/VR, and unconstrained in-the-wild video.


\bibliographystyle{IEEEtran}
\bibliography{IEEEabrv,IEEE_resources}

\end{document}